\documentclass[letterpaper, 10 pt, conference]{ieeeconf}  % Comment this line out if you need a4paper

\IEEEoverridecommandlockouts                              % This command is only needed if 
\usepackage{graphicx}
\usepackage{subcaption}
\usepackage{booktabs}
\usepackage{multirow}
\usepackage{amsfonts}
\usepackage{amsmath}
\usepackage{xcolor}
\usepackage[table,xcdraw]{xcolor}
\usepackage[colorlinks=true, linkcolor=red]{hyperref}
\usepackage{bm}
\usepackage{tabularx}

\usepackage{siunitx}

\usepackage[ruled,noline]{algorithm2e} % 漂亮的算法环境

\DeclareMathOperator*{\argmin}{arg\,min}

\usepackage{bbding}

\let\labelindent\relax
\usepackage{enumitem}

\usepackage{tikz}
\usepackage{pgfplots}
\pgfplotsset{compat=1.18}

\usepackage{pifont}

\title{\LARGE \bf MapRF: Weakly Supervised Online HD Map Construction \\ via NeRF-Guided Self-Training
}

\author{
Hongyu Lyu\textsuperscript{1}, Thomas Monninger\textsuperscript{2}, Julie Stephany Berrio Perez\textsuperscript{1}, Mao Shan\textsuperscript{1}, \\ Zhenxing Ming\textsuperscript{1}, Stewart Worrall\textsuperscript{1} % <-this % stops a space
\thanks{\textsuperscript{1}The University of Sydney, Australian Centre for Robotics, Sydney, Australia. Emails: \texttt{\small \{h.lyu, j.berrio, m.shan, d.ming, s.worrall\}@acfr.usyd.edu.au}}
\thanks{\textsuperscript{2}Mercedes-Benz Research \& Development North America, San Jose, CA, USA. Email: \texttt{\small thomas.monninger@mercedes-benz.com}}
}

\begin{document}

\vspace{-15mm}   % adjust this value
\maketitle
\thispagestyle{empty}
\pagestyle{empty}

%%%%%%%%%%%%%%%%%%%%%%%%%%%%%%%%%%%%%%%%%%%%%%%%%%%%%%%%%%%%%%%%%%%%%%%%%%%%%%%%
\begin{abstract}

Autonomous driving systems benefit from high-definition (HD) maps that provide critical information about road infrastructure. The online construction of HD maps offers a scalable approach to generate local maps from on-board sensors. However, existing methods typically rely on costly 3D map annotations for training, which limits their generalization and scalability across diverse driving environments. In this work, we propose MapRF, a weakly supervised framework that learns to construct 3D maps using only 2D image labels. To generate high-quality pseudo labels, we introduce a novel Neural Radiance Fields (NeRF) module conditioned on map predictions, which reconstructs view-consistent 3D geometry and semantics. These pseudo labels are then iteratively used to refine the map network in a self-training manner, enabling progressive improvement without additional supervision. Furthermore, to mitigate error accumulation during self-training, we propose a Map-to-Ray Matching strategy that aligns map predictions with camera rays derived from 2D labels. Extensive experiments on the Argoverse~2 and nuScenes datasets demonstrate that MapRF achieves performance comparable to fully supervised methods, attaining around 75\% of the baseline while surpassing several approaches using only 2D labels. This highlights the potential of MapRF to enable scalable and cost-effective online HD map construction for autonomous driving.

\end{abstract}

%%%%%%%%%%%%%%%%%%%%%%%%%%%%%%%%%%%%%%%%%%%%%%%%%%%%%%%%%%%%%%%%%%%%%%%%%%%%%%%%

\section{Introduction}

% macro background
High-Definition (HD) maps are a critical component of autonomous driving systems, providing high-precision prior information about road infrastructure to support downstream modules~\cite{lyu2025online}. They typically include map elements such as lane dividers, road boundaries, and pedestrian crossings, capturing their geometry, semantics, and topology~\cite{elghazaly2023high, monninger2023semantic}. Traditionally, HD maps are constructed offline from multimodal data collected by survey vehicles, processed through SLAM-based pipelines~\cite{zhang2014loam, shan2018lego}, and refined with manual annotation. Although accurate, this process is expensive and difficult to maintain in dynamic road environments~\cite{Berrio2020LongTermMM}. Consequently, the construction of HD maps online has attracted increasing attention, with the intention of generating local maps at runtime from sensor data on-board~\cite{liu2023vectormapnet, liao2024maptrv2, monninger2025augmapnet, monninger2025mapdiffusion}.

% paper motivation
Despite recent advances in online HD map construction, a significant challenge in this domain is the reliance on extensive ground-truth map labels for supervised learning~\cite{lowens2025pseudomaptrainer}. Specifically, to train a model that generates local maps from on-board sensor data, one typically first needs to identify map elements of interest in images and then annotate them in 3D LiDAR point clouds with cross-modal validation~\cite{lyu2025online}. This annotation process is expensive and labor-intensive, and the resulting models often generalize poorly when the training and test data distributions diverge, limiting the scalability of autonomous driving systems~\cite{ranganatha2024semvecnet}. In contrast, 2D labels, such as lane markings, are much easier to obtain, as they require only annotation of the image domain and have been collected extensively over the past decade~\cite{ai2022ws}.

\begin{figure}[t!] % The asterisk (*) makes the figure span both columns
  \centering
  \includegraphics[width=\columnwidth]{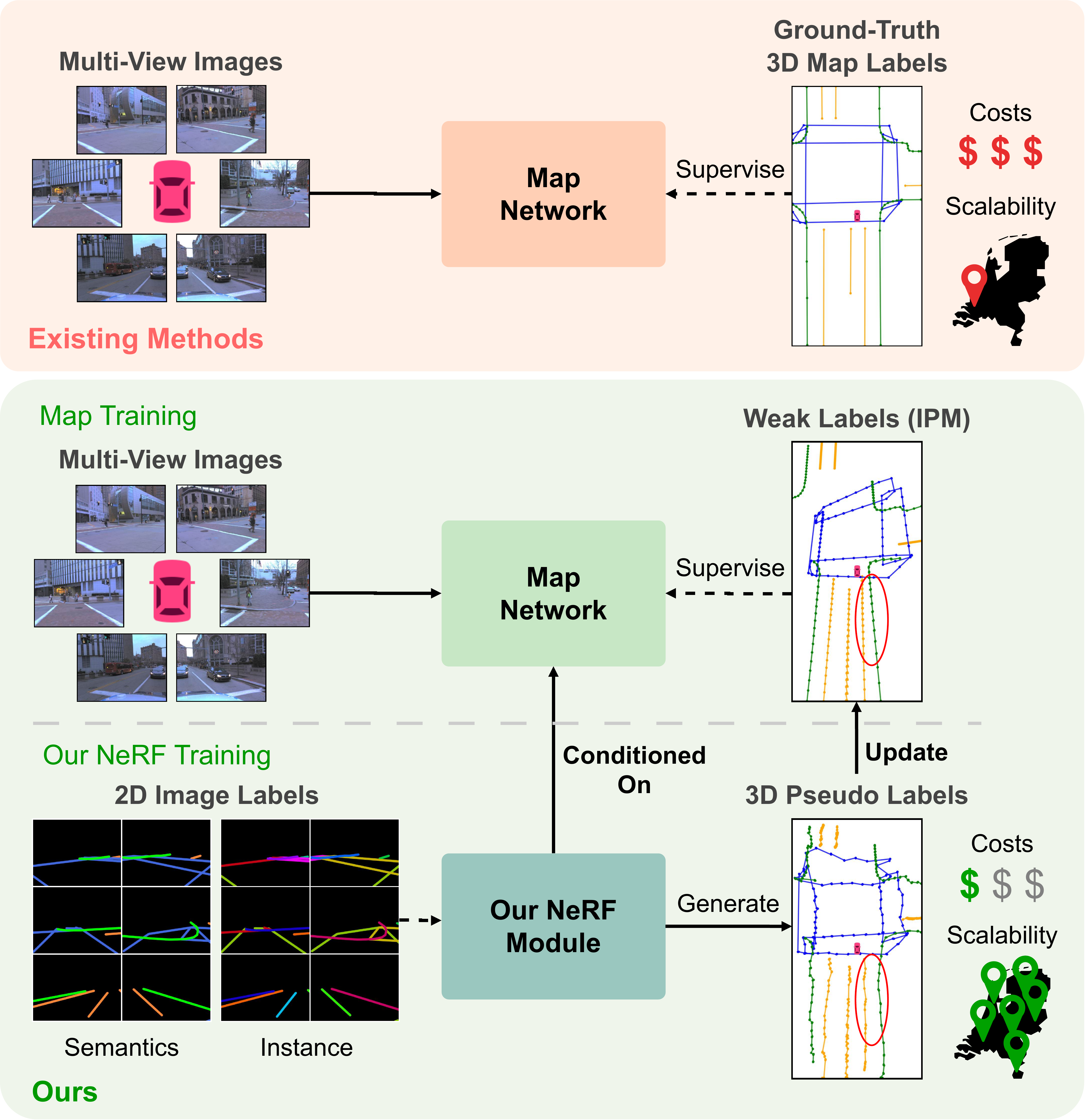}
  \caption{\small \textbf{Motivation for MapRF.} Compared to existing methods, MapRF learns from accessible 2D image labels to construct 3D HD maps online. We generate pseudo labels through the proposed NeRF module and use them for self-training. This design reduces data collection and annotation costs, thereby improving scalability.}
  \label{fig: motivation}
  \vspace{-5mm}
\end{figure}

These observations motivate the question: \textit{Can we learn a model from 2D image labels that can construct 3D maps online?} To this end, we propose MapRF, a weakly supervised framework that eliminates the need for costly 3D map labels during training (see Fig.~\ref{fig: motivation}). Specifically, we generate pseudo labels based on local map reconstruction using Neural Radiance Fields (NeRF)~\cite{mildenhall2021nerf}. We introduce a Map-Conditioned NeRF (MC-NeRF) that models local maps by conditioning on map predictions and optimizing with multi-view image labels. Since this representation encodes not only semantic and instance properties, but also geometry, we can derive pseudo labels from the density field. These pseudo labels can serve as a basis for self-training, allowing the map model to be progressively improved.
% MapRF assumes the availability of 2D semantic and instance annotations in the image domain. In the absence of such annotations, the framework cannot be directly applied, as both the initial IPM-based supervision and the NeRF-guided self-training rely on image-level semantic cues. Exploring fully unsupervised or foundation-model–derived alternatives to 2D labels is left for future work.

Although self-training provides an effective way to leverage pseudo labels for iterative improvement, it is also prone to concept drift~\cite{you2022learning}. Since the model relies repeatedly on its own predictions to derive supervision signals, systematic errors can accumulate over iterations and gradually bias the learning process. To address this, we introduce a Map-to-Ray Matching (MRM) strategy to enhance pseudo-label generation. In particular, MRM enforces consistent alignment between map predictions and 2D labels by associating predicted map elements with semantic ray groups. 

Our main contributions can be summarized as follows:

\begin{itemize}[leftmargin=*]\setlength{\itemsep}{0.2em}
    \item  We present a weakly supervised framework for online HD map construction, which learns from 2D image labels to reduce the reliance on expensive 3D map annotations.

    \item We propose a novel Map-Conditioned NeRF that jointly models geometric, semantic, and instance information to generate high-quality pseudo labels for map learning.

    \item We introduce a Map-to-Ray Matching strategy that mitigates concept drift during self-training by enforcing consistent alignment between map predictions and 2D labels.

    \item Extensive experiments on two large-scale datasets demonstrate competitive performance and validate the effectiveness of the proposed framework.
\end{itemize}  

The remainder of this paper is organized as follows. Section~\ref{sec:related_work} reviews related work. 
Section~\ref{sec:method} describes the proposed MapRF framework. 
Section~\ref{sec:experiments} presents experimental results and analysis. 
Section~\ref{sec:conclusion} concludes the paper.

\section{Related Work}  \label{sec:related_work}

% Existing methods~\cite{liao2024maptrv2, monninger2025augmapnet, mink2024lmt} can be broadly classified into two categories based on their output representations. Raster-based methods~\cite{li2022hdmapnet, peng2023bevsegformer, dong2024superfusion} formulate the task as semantic segmentation in the bird’s-eye view (BEV) space, generating rasterized maps that require post-processing to extract vectorized elements. In contrast, vector-based methods~\cite{liao2022maptr, liu2023vectormapnet, qiao2023end} directly predict map elements as point sets or polylines, typically employing DETR-style~\cite{carion2020end} architectures for end-to-end vectorized map construction.

% Semi-BEVSeg~\cite{zhu2024semi} introduces a conjoint rotation augmentation and enforces consistency in both predictions and BEV features. PCT~\cite{ishikawa2024pct} proposes a camera dropout augmentation and leverages perspective-view pseudo labels from pretrained models. Lilja et al.~\cite{lilja2025exploring} further explore temporal fusion by aggregating teacher predictions and BEV features across nearby frames.

% SkyEye~\cite{gosala2023skyeye} enforces spatiotemporal consistency across frames and generates BEV pseudo labels using self-supervised depth. RendBEV \cite{monteagudo2025rendbev} employs volumetric rendering to learn directly from 2D semantic supervision.

\noindent \textbf{Online HD Map Construction} has emerged to reduce the cost of creating and maintaining HD maps by generating local maps at runtime from on-board sensor data. HDMapNet~\cite{li2022hdmapnet} pioneers convolutional networks to segment map elements, followed by post-processing to obtain vectorized maps. VectorMapNet~\cite{liu2023vectormapnet} presents the first end-to-end framework that predicts vectorized map elements as point sequences with an autoregressive model. Subsequent work explores various map representations, including permutation-equivalent~\cite{liao2022maptr, liao2024maptrv2}, graph-based~\cite{shin2025instagram}, piecewise Bézier curve~\cite{qiao2023end}, pivot-based~\cite{ding2023pivotnet}, Douglas-Peucker~\cite{liu2024compact}, and transformation-invariant~\cite{zhang2024onlinegeometry}. MapVR~\cite{zhang2024online} introduces differentiable rasterization to provide fine-grained geometric supervision. StreamMapNet~\cite{yuan2024streammapnet} and SQD-MapNet~\cite{wang2024stream} exploit temporal information with temporal fusion and query denoising. MapQR~\cite{liu2024leveraging} proposes a scatter-and-gather query design to enhance the interaction of points and map elements. Recently, efforts have been made to reduce the reliance on expensive map annotations. PseudoMapTrainer~\cite{lowens2025pseudomaptrainer} employs a pretrained segmentor and road surface reconstruction to generate pseudo labels for map learning. To the best of our knowledge, our framework is the first to learn from 2D image labels and construct 3D HD maps online.

\vspace{0.1cm}
\noindent \textbf{Label-Efficient BEV Map Segmentation} seeks to generate rasterized bird’s-eye-view (BEV) semantic maps while reducing the reliance on costly BEV map labels. S2G2~\cite{gao2022s2g2} pioneers a semi-supervised framework that uses unlabeled data through a teacher–student consistency scheme. Subsequent work further advances this direction through data augmentations~\cite{zhu2024semi, ishikawa2024pct}, perspective-view cues~\cite{ishikawa2024pct}, and teacher temporal fusion~\cite{lilja2025exploring}. Another line of research explores self-supervised pretraining to learn transferable BEV features from unlabeled data. LetsMap~\cite{gosala2024letsmap} employs implicit fields for geometric modeling and a temporal masked autoencoder for representation learning. OccFeat~\cite{sirko2024occfeat} proposes pretraining through occupancy reconstruction and feature distillation from image foundation models. More recently, methods have used 2D semantic labels to train BEV networks, either by generating BEV pseudo labels~\cite{gosala2023skyeye} or directly learning from 2D supervision~\cite{monteagudo2025rendbev}. Unlike these methods, our work constructs vectorized 3D HD maps with explicit geometric and semantic representations.

\vspace{0.1cm}
\noindent \textbf{3D Reconstruction} aims to recover both the geometry and appearance of the scene from multi-view images. Traditional pipelines, typically based on Structure-from-Motion~\cite{schonberger2016structure} and Multi-View Stereo~\cite{schonberger2016pixelwise}, rely on accurate feature correspondences and often degrade in low-texture or reflective scenes. Neural Radiance Fields (NeRF)~\cite{mildenhall2021nerf} addresses these limitations by representing scenes as continuous functions parameterized by neural networks. PixelNeRF~\cite{yu2021pixelnerf} extends NeRF by conditioning on pixel-aligned image features, allowing generalizable prediction to unseen scenes. PointNeRF~\cite{xu2022point} and PointNeRF++~\cite{sun2024pointnerf++} integrate point clouds as explicit geometric priors in the representation, achieving faster convergence and more accurate reconstruction. Beyond general scene reconstruction, mesh-based~\cite{mei2024rome, wu2024emie} and Gaussian-based~\cite{feng2024rogs} representations have been explored for road surface reconstruction. Unlike these methods, our work proposes a novel NeRF design tailored for online HD map construction, which jointly models geometry, semantics, and instances to generate pseudo labels that replace map labels.

% \vspace{0.1cm}
% \noindent \textbf{Lane Detection} aims to identify the position and shape of lane markings and can be regarded as a subtask of online HD map construction. Early 2D-based approaches detect lane markings from front-view images, including mask-based~\cite{lee2017vpgnet, pan2018spatial}, grid-based~\cite{qin2020ultra, liu2021condlanenet}, keypoint-based~\cite{ko2021key, qu2021focus}, line-anchor-based~\cite{li2019line, tabelini2021keep}, and curve-based approaches~\cite{ liu2021end, li2023pga}. To project 2D lane coordinates into 3D space, inverse perspective mapping (IPM)~\cite{mallot1991inverse} or depth estimation~\cite{yan2022once} is commonly employed. However, IPM assumes a flat surface and therefore suffers from distortions on nonplanar roads. Recent advances directly predict 3D lane geometry, either through BEV-based methods~\cite{garnett20193d, chen2022persformer} that take advantage of view transformation and height information, or non-BEV-based methods~\cite{bai2023curveformer, huang2023anchor3dlane} that model lanes directly in 3D coordinates. Unlike lane detection, our work uses accessible 2D labels to construct local maps around the vehicle, which include a richer set of map elements.

In summary, existing methods for online HD map construction typically rely on costly map annotations, while 3D reconstruction approaches primarily focus on general scenes rather than map semantics. To bridge this gap, we propose MapRF, a NeRF-guided weakly supervised framework that constructs 3D vectorized maps efficiently and accurately.

% In contrast, our approach processes the pseudo and real 3D
% point clouds, which come from surround-view cameras and
% lidar, respectively, under cylindrical coordinates for better
% representation and to preserve more fine-grained geometry
% information that leads to better performance.

\section{Method} \label{sec:method}

\begin{figure*}[t!] % The asterisk (*) makes the figure span both columns
  \centering
  \vspace{2mm}
  \includegraphics[width=0.9\textwidth]{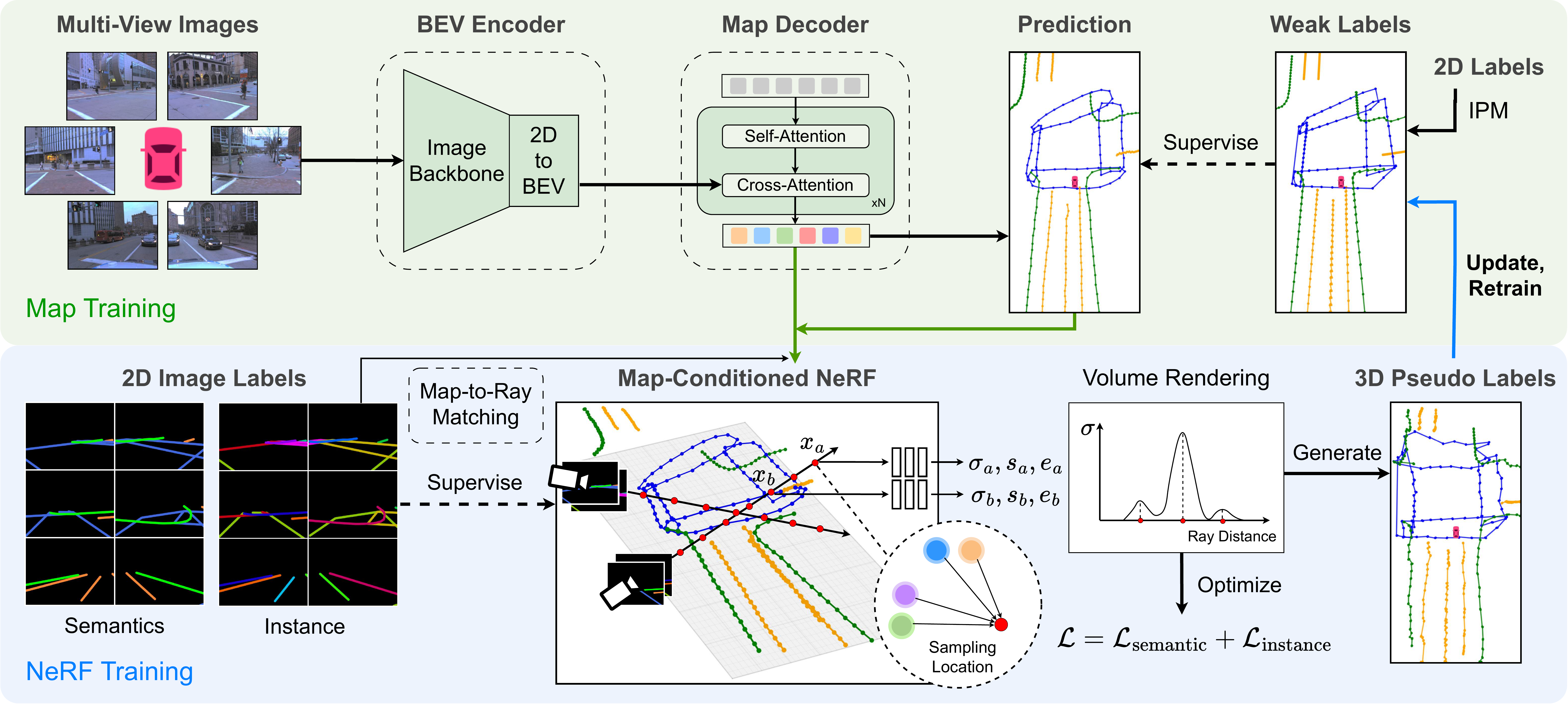}
  \caption{\small \textbf{Overall framework of MapRF.} The framework learns to construct 3D HD maps online using only 2D image annotations. We first train an initial map model with weak labels generated via IPM. We then optimize a Map-Conditioned NeRF (MC-NeRF) with multi-view labels to generate pseudo labels. We iteratively retrain the map model with pseudo labels and re-optimize MC-NeRF, forming a self-training loop. We further employ a Map-to-Ray Matching (MRM) strategy to mitigate the risk of concept drift during self-training.}
  \label{fig: framework}
  \vspace{-4mm}
\end{figure*}

\noindent \textbf{Problem Setup.} We aim to develop a model that constructs local maps online from multi-view images by predicting map elements, each defined by a semantic class and a sequence of 3D points. We consider a weakly supervised setting in which the model is trained using only 2D semantic and instance annotations in the image domain, without access to ground-truth 3D map labels. The data can be collected by a vehicle equipped with synchronized sensors, including surrounding cameras for multi-view imaging and kinematic sensors for relative pose estimation. This setting is practical and cost-effective, as 2D image annotations are easier to obtain than 3D map labels and do not require range sensors during data collection. Our framework assumes the availability of temporally consistent semantic and instance attributes in 2D labels, which provide the necessary supervision signal for training.

\vspace{0.1cm}
\noindent \textbf{Overview.} We propose MapRF, a weakly supervised framework that uses 2D image labels for map learning.  As shown in Fig.~\ref{fig: framework}, the framework first learns an initial map model from weak labels generated via IPM with heuristic constraints. Building on this model, we introduce a Map-Conditioned NeRF (MC-NeRF) that generates pseudo labels for self-training by conditioning on map predictions and optimizing with multi-view image labels. Specifically, the MC-NeRF enforces geometric consistency through multi-view reconstruction, gradually refining the structure and location of map elements to guide the self-training process. To mitigate the risk of concept drift during self-training, we further propose a Map-to-Ray Matching (MRM) strategy that ensures consistent alignment between map predictions and 2D labels. The general procedure is summarized in Algorithm~\ref{alg:maprf}.

\subsection{Learning an Initial Map Model from Weak Labels}

Self-training methods~\cite{you2022learning} typically rely on an initial set of labels that provide a starting point for iterative refinement. In the absence of ground-truth map annotations, we derive weak labels from 2D annotations to train an initial map model. Specifically, we employ inverse perspective mapping (IPM)~\cite{mallot1991inverse} to project semantic and instance labels onto the ground plane and group points by instance identity. However, IPM relies on a globally planar ground assumption, which may introduce geometric distortions in scenes with elevation changes. To mitigate these distortions, we relax the single-plane constraint by assigning an elevation parameter to each projected instance. This parameter is estimated by selecting the elevation that maximizes cross-frame reprojection consistency. The resulting weak labels are shown in Fig.~\ref{fig: weak_labels}.

% IPM assumes a perfectly planar road surface, which does not always hold in real-world environments (see Fig.~\ref{fig: weak_labels}). To alleviate the resulting distortions, we consider heuristic constraints during the projection process, assigning each map element a plausible height inferred from geometric cues. 

% Specifically, we constrain each projected instance with class-dependent height priors and enforce smooth height variation along the instance trajectory to suppress projection artifacts caused by road slopes. These heuristics reduce severe distortions but cannot fully recover non-planar geometry, as in Fig. \ref{fig: weak_labels}.

\subsection{Generating Pseudo Labels with Map-Conditioned NeRF}

NeRF~\cite{mildenhall2021nerf} offers a powerful representation for 3D scene reconstruction by jointly modeling geometry and appearance. However, pseudo labels for map learning require more than geometric fidelity; they must also capture semantic categories and instance properties of map elements. To address this, we propose the MC-NeRF module, which conditions on map predictions and reconstructs map elements to generate high-quality pseudo labels. We start with some preliminaries.

\begin{algorithm}[t]
\small

\caption{Pseudocode of MapRF Self-Training}
\label{alg:maprf}
\DontPrintSemicolon
\SetKwInOut{Input}{Input}
\SetKwInOut{Output}{Output}

\SetKw{KwDo}{do}

\SetKwFor{For}{for}{}{}

\Input{$\{I_t\}$ multi-view images with camera poses\\
$\{L_t\}$ 2D semantic and instance labels\\
$R_{\max}$ maximum self-training (ST) rounds}
\Output{$M_{R_{\max}}$ the map model after $R_{\max}$ rounds}

$\mathcal{P}_0 \leftarrow \text{IPM\_with\_heuristics}(\{L_t\})$\;
$M_0 \leftarrow \text{train\_map}(\{I_t\},\, \mathcal{P}_0)$  \hspace{1.1cm} $\triangleright$ \text{0-th round training}  \; 

\For{$r \leftarrow 1$ \KwTo $R_{\max}$ \KwDo \hspace{1.15cm} $\triangleright$ \textnormal{$r$-th round self-training}  } 
{ 
  $\hat{\mathcal{Y}}_r \leftarrow \text{get\_prediction}(M_{r-1},\, \{I_t\})$\;
  $\hat{\mathcal{Y}}_r \leftarrow \text{perform\_MRM}(\hat{\mathcal{Y}}_r,\, \{L_t\})$\;
  $N_r \leftarrow \text{train\_NeRF}(\hat{\mathcal{Y}}_r,\, \{L_t\})$ \;
  $\mathcal{P}_r \leftarrow \text{generate\_pseudo\_labels}(N_r, \hat{\mathcal{Y}}_r,\, \{L_t\})$\;
  $M_r \leftarrow \text{train\_map}(\{I_t\},\, \mathcal{P}_r)$  \; 
}
\textbf{end for}

\end{algorithm}

% \vspace{0.1cm}
\noindent \textbf{NeRF and Volume Rendering.}
NeRF~\cite{mildenhall2021nerf} represents a 3D scene as a continuous function parameterized by a multilayer perceptron (MLP). 
For any 3D position $x \in \mathbb{R}^3$ and viewing direction $d \in \mathbb{S}^2$, the network outputs a density $\sigma(x)$ and a color $c(x, d)$. 
For each pixel, a ray is cast into the scene and discretized into sampling points $\{x_i\}_{i=1}^{M}$. 
The pixel color is obtained by accumulating contributions from all samples:
\begin{equation}
C = \sum_{i=1}^{M} \tau_i \bigl(1 - \exp(-\sigma_i \,\delta_i)\bigr)\, c_i.
\end{equation}
where $\delta_i$ is the distance between adjacent samples and $\tau_i$ denotes the accumulated transmittance up to the $i$-th sample.

\vspace{0.1cm}
\noindent \textbf{Map-Conditioned NeRF.} We represent a map prediction as a set of neural points $\mathcal{P} = \{(p_i,\, f_i)\mid i = 1, \ldots, N\}$, where each point $i$ is located at $p_i$ and associated with a feature vector $f_i$ that encodes the local context of a map element. We regress semantic and instance properties from this representation.

Given any 3D location $x$, we query $K$ neighboring neural points around it. Our MC-NeRF can be abstracted as a neural module that regresses volume density $\sigma$, semantic logits $s$, and instance embedding $e$ at the sampling location $x$ from its neighboring neural points, as:
\begin{equation}
(\sigma, s, e) = \text{MC-NeRF}\bigl(x,\,  \{(p_i,\,  f_i)\}_{i=1}^K \bigr).
\label{eq: mc-nerf}
\end{equation}
We employ a PointNet-style~\cite{qi2017pointnet} architecture with multiple sub-MLPs for this regression. Overall, we first transform map predictions into neural point representations and then aggregate the multi-point information to obtain the outputs. 
Conditioned on map predictions, MC-NeRF learns structural patterns that are consistent and transferable across scenes, enabling generalization without per-scene optimization. 
% Each MLP consists of three fully connected layers with ReLU activations and hidden dimensions of 64. The aggregation follows a max-pooling operation as in PointNet~\cite{qi2017pointnet} . Unless stated otherwise, all MLPs share this configuration

\vspace{0.1cm}
\noindent \textbf{Map Prediction Processing.} For each predicted point, we employ two MLPs to obtain its neural point representation: $F_1$ refines the raw position $q_i$ using the associated instance feature $f_{\text{ins}}$, and $F_2$ produces the point feature from $f_{\text{ins}}$ and the refined position.
\begin{equation}
    p_i = q_i + F_1(f_{\text{ins}}), \quad 
    f_i = F_2(f_{\text{ins}},\, p_i). 
\end{equation}
We then use an additional MLP $F_3$ to generate a new feature vector $f_{i,x}$ relative to the sampling location $x$:
\begin{equation}
    f_{i,x} = F_3(f_i,\,x - p_i).
\end{equation}
Here, the feature $f_i$ encodes the local context of the map element around $p_i$. Leveraging the relative position $x - p_i$, the MLP $F_3$ generates $f_{i,x}$ in the local frame of the neural point, resulting in translation invariance and better generalization.

\vspace{0.1cm}
\noindent \textbf{Sampling Feature Encoding.} For each sampling location $x$, we use inverse distance weighting to aggregate features $f_{i,x}$ from its $K$ neighboring neural points to obtain $\tilde{f}_x$. Then, an MLP $F_4$ integrates $\tilde{f}_x$ with $x$ to obtain the feature $f_x$:
\begin{equation}
\tilde{f}_x=\sum_{i=1}^K \frac{w_i}{\sum w_i}\,f_{i,x},
\quad w_i=\frac{1}{\|x-p_i\|^2}.
\end{equation}
\begin{equation}
    f_x = F_4(\tilde{f}_x,\, x).
\end{equation}
Here, inverse distance weighting provides a natural way to aggregate features from neural points.  Points closer to $x$ are assigned higher weights, while farther ones contribute less. The squared distance emphasizes locality for large scenes.

\vspace{0.1cm}
\noindent \textbf{Field Prediction and Rendering.} Given the encoded feature $f_x$ at $x$, we use three MLPs to predict volume density $\sigma_x$, semantic logits $s_x$, and instance embedding $e_x$:
\begin{equation}
    \sigma_x = F_5(f_x), \quad 
    s_x = F_6(f_x), \quad 
    e_x = F_7(f_x).
\end{equation}
For each labeled pixel, a ray is cast into the scene and discretized into sampling points $\{x_j\}_{j=1}^{M}$. We then accumulate the predictions along the ray to produce the per-pixel outputs:
\begin{equation}
S = \sum_{j=1}^{M} \tau_j \bigl(1 - \exp(-\sigma_j \,\delta_j)\bigr)\, s_j.
\end{equation}
\begin{equation}
E = \sum_{j=1}^{M} \tau_j \bigl(1 - \exp(-\sigma_j \,\delta_j)\bigr)\, e_j.
\end{equation}
where $\delta_i$ is the sample interval and $\tau_i$ is the transmittance.

% Here, the shared feature $f_x$ serves as a unified representation of the map elements around $x$. Since these properties are inherently view-invariant, the predictions are independent of the viewing direction. 

\vspace{0.1cm}
\noindent \textbf{Optimization.} We optimize MC-NeRF by supervising the rendered outputs with 2D image annotations. Specifically, we employ a focal loss~\cite{lin2017focal} for the semantic logits and a discriminative loss~\cite{de2017semantic} for the instance embeddings:
\begin{equation}
\mathcal{L} = \lambda_s \mathcal{L}_\text{semantic} + \lambda_i \mathcal{L}_\text{instance}.
\end{equation}
Here, the $\mathcal{L}_\text{instance}$ structures the embedding space by enforcing intra-instance compactness and inter-instance separation. The underlying geometry is implicitly optimized to be consistent across views, which supports pseudo-label generation.

\begin{figure}[t!] % The asterisk (*) makes the figure span both columns
  \centering
  \vspace{2mm}
  \includegraphics[width=\columnwidth]{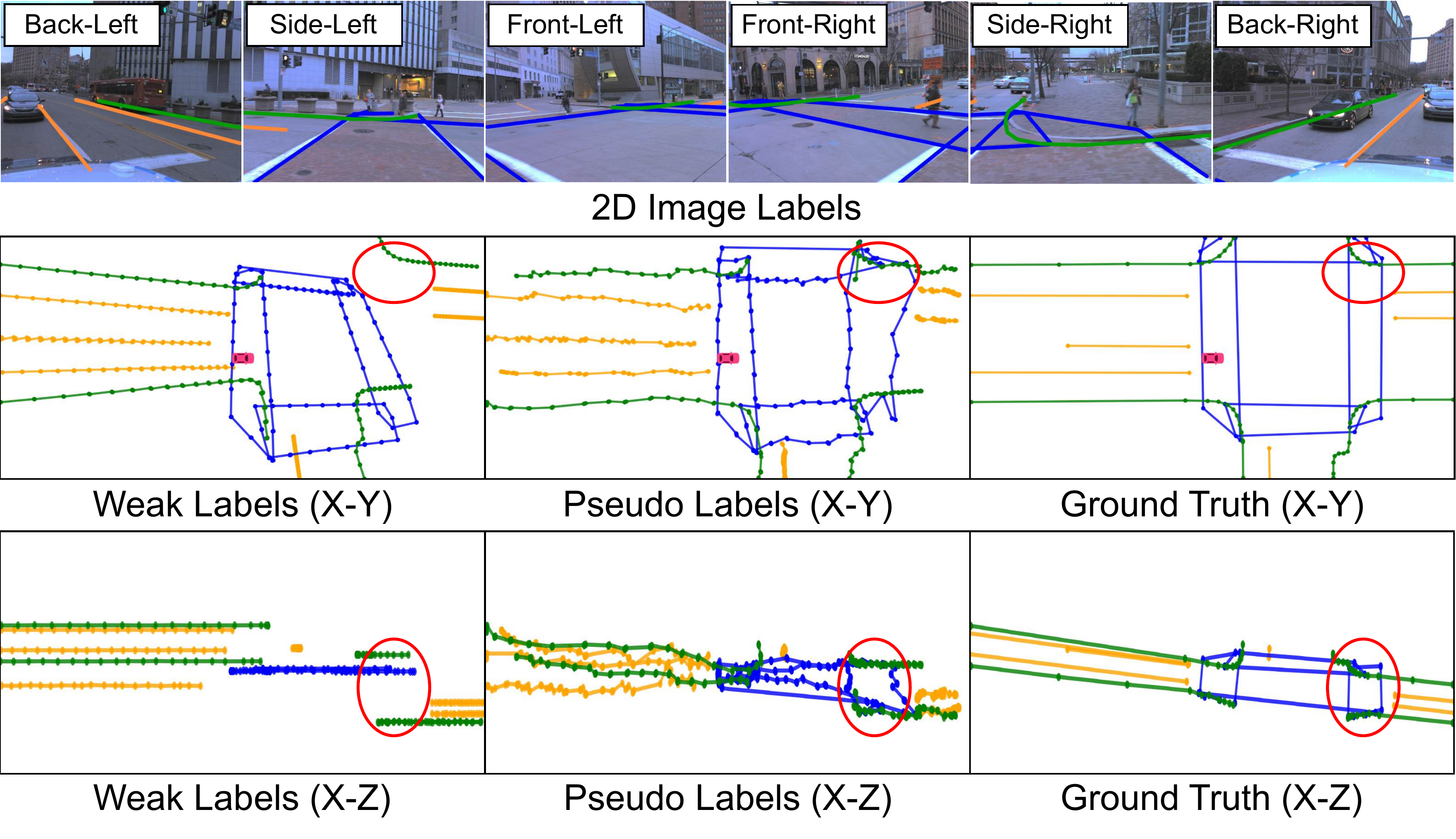}
  \caption{\small \textbf{Weak vs. pseudo labels.} The planar ground assumption in IPM introduces geometric distortions in scenes with elevation changes. Instance-level elevation estimation alleviates these artifacts, while our pseudo labels further improve geometric coherence.}
   \vspace{-4mm}
  \label{fig: weak_labels}
\end{figure}

\vspace{0.1cm}
\noindent \textbf{Pseudo-Label Generation.} We use MC-NeRF to generate pseudo labels from 2D image annotations. For each labeled pixel, a ray is cast into the scene, and the sampling point with the maximum predicted density is selected. These points are then grouped by instance identity and organized into map elements, forming pseudo labels for map learning. An example is shown in Fig.~\ref{fig: weak_labels}.
Compared to weak labels, the pseudo labels exhibit reduced geometric distortion, as multi-view consistency provides stronger geometric constraints during optimization. However, they may appear noisier due to uncertainty in density prediction and discrete ray sampling, which can introduce slight cross-ray inconsistencies within an instance. This noise has limited impact on map learning, as the network prioritizes consistent geometric patterns over minor inconsistencies.

% Compared to IPM-based weak labels, the pseudo labels generated by MC-NeRF has less distortion because they are optimized through multi-view consistency rather than a single planar assumption. However, they may appear noisier due to uncertainty in density estimation and ray sampling, especially in regions with limited visual evidence. This noise is mitigated during self-training by instance aggregation and Map-to-Ray Matching.

\subsection{Bootstrapping the Map Model with Self-Training}

\begin{table*}[t!]
\centering
\vspace{2mm}
\small
\caption{\small \textbf{Comparison with state-of-the-art methods on the Argoverse 2 validation set.} Results are reported for BEV and 3D maps within a perception range of $\SI{60}{\meter}\times\SI{30}{\meter}\times\SI{8}{\meter}$. “Map GT” indicates whether ground-truth HD map annotations are used for training. $^*$~denotes results reproduced from public code; “–” means results not reported. Other fully supervised results are taken from \cite{liao2024maptrv2, liu2024leveraging}.}

\begin{tabular}{l|c|cccc|cccc}
\toprule
\multicolumn{1}{c|}{\multirow{2}{*}{Method}}
  & \multirow{2}{*}{Map GT}
  & \multicolumn{4}{c|}{BEV}
  & \multicolumn{4}{c}{3D} \\
\cmidrule(lr){3-6} \cmidrule(lr){7-10}
  & 
  & AP$_{\text{div}}$ & AP$_{\text{ped}}$ & AP$_{\text{bou}}$ & mAP$_{\text{1}}$
  & AP$_{\text{div}}$ & AP$_{\text{ped}}$ & AP$_{\text{bou}}$ & mAP$_{\text{2}}$ \\
\midrule
HDMapNet \cite{li2022hdmapnet}    & \ding{51} & 5.7 & 13.1 & 37.6 & 18.8 & – & – & – & – \\
VectorMapNet \cite{liu2023vectormapnet}    & \ding{51} & 36.1 & 38.3 & 39.2 & 37.9 & 35.0 & 36.5 & 36.2 & 35.8 \\
MapTR \cite{liao2022maptr}      & \ding{51} & 58.7 & 55.4 & 59.1 & 57.8 & 61.6$^{*}$ & 54.9$^{*}$ & 59.2$^{*}$ & 58.6$^{*}$ \\
MapTRv2 \cite{liao2024maptrv2}   & \ding{51} & 72.1  & 62.9 & 67.1 & 67.4 & 69.1 & 59.8 & 65.3 & 64.7 \\
MapQR \cite{liu2024leveraging}             & \ding{51} & 72.3 & 64.3 &  68.1 & \textbf{68.2} & 71.2 & 60.1 & 66.2 & \textbf{65.9} \\
\midrule
MapQR \cite{liu2024leveraging} + IPM \cite{mallot1991inverse}       & \ding{55} & 55.8 & 31.2 & 40.3 & 42.4 & 54.4 & 29.7 & 38.6 & 40.9 \\
% PseudoMapTrainer \cite{lowens2025pseudomaptrainer}  & \ding{55} & – & – & – & – & – & – & – & – \\
MapRF (Ours)                               & \ding{55} & 59.6 & 41.8 & 51.0 & \textbf{50.8} & 58.1 & 40.5 & 49.8 & \textbf{49.5} \\
\bottomrule
\end{tabular}

\label{tab:av2_results} 
\vspace{-2mm}
% \vspace{-5mm}
\end{table*}

Self-training bootstraps models by generating pseudo labels from their own predictions for iterative retraining~\cite{you2022learning}. In our framework, the MC-NeRF can generate pseudo labels \textit{of higher quality} than the initial weak labels. By enforcing multi-view consistency, it preserves geometric fidelity during local map reconstruction, mitigating distortions from IPM and recovering the fine-grained structures of map elements. Leveraging these pseudo labels, we iteratively retrain the map model and re-optimize MC-NeRF, progressively improving both map predictions and supervision signals. This process will eventually converge when the pseudo labels stabilize and the map model no longer improves.

While self-training can iteratively improve the map model, it remains susceptible to concept drift, where systematic errors in predictions are propagated through pseudo labels and accumulate over iterations, biasing the learning process. To address this, we propose a Map-to-Ray Matching strategy that aligns map predictions with 2D annotations and selects reliable instances as conditional inputs for MC-NeRF.

\vspace{0.1cm}
\noindent \textbf{Map-to-Ray Matching.} For each scene, we represent the 2D labels as a set of semantic ray groups $\mathcal{R} = \{ (s_m, R_m^{\Gamma}) \}_{m=1}^{N_\text{gt}} $, where each map element $m$ is associated with a class label $s_m$ and a ray group $R_m$ from its labeled pixels. We further augment $R_m$ with equivalent permutations $\Gamma$ to account for ordering ambiguity, inspired by~\cite{liao2024maptrv2}.

Given a set of predicted map elements, we establish correspondences with $\mathcal{R}$ and retain only those matched instances that are geometrically consistent. The optimal assignment $\hat{\pi}$ is obtained by minimizing the total matching cost:
\begin{equation}
\hat{\pi} = \argmin_{\pi \in \Pi_N}
\sum_{(n,m)\in\pi}
\Big[
\mathcal{L}_{\text{cls}}(\hat{s}_n, s_m)
+
\mathcal{L}_{\text{geo}}(\hat{P}_n, R_m^{\Gamma})
\Big].
\end{equation}
Here, $\mathcal{L}_{\text{cls}}$ is a focal cost~\cite{lin2017focal} between the predicted class score $\hat{s}_n$ and the label $s_m$, while $\mathcal{L}_{\text{geo}}$ measures geometric similarity between the predicted point sequence $\hat{P}_n$ and the ray group $R_m^{\Gamma}$. To compute $\mathcal{L}_{\text{geo}}$, we search for the optimal permutation $\hat{\gamma} \in \Gamma$ that minimizes the point-to-ray distance:
\begin{equation}
\hat{\gamma} = \argmin_{\gamma \in \Gamma}
\frac{1}{|R_m|}
\sum_{j=1}^{|R_m|}
D_{\text{L1}}(\hat{p}_j, r_{\gamma(j)}).
\end{equation}
where $D_{\text{L1}}$ denotes the L1 distance between a predicted point $\hat{p}_j$
and its projection onto the ray $r_{\gamma(j)}$. The overall assignment $\hat{\pi}$ is solved using the Hungarian algorithm.

\section{Experiments} \label{sec:experiments}

% We evaluate the proposed MapRF by comparing it with several baselines on the Argoverse~2 and nuScenes datasets. To further analyze its components, we also conduct extensive ablation studies on Argoverse~2 in Sec.~\ref{sec: ablation}.

\subsection{Experimental Setup}

% Argoverse~2 contains 1,000 logs, each 15 seconds long, collected from six U.S. cities. Each log includes RGB images from seven 20Hz cameras and log-level 3D vector maps annotated at 10Hz. NuScenes includes 1,000 scenes, each 20 seconds long, collected from Boston and Singapore. It provides RGB images from six 12Hz cameras and city-scale BEV vector maps annotated at 2Hz. As nuScenes provides only BEV vector maps, we extract height information from point clouds to recover 3D geometry \textcolor{red}{TODO}. Notably, MapRF is trained using only 2D image labels, making the task more challenging than baselines that use full annotations.

% \vspace{0.1cm} 
\noindent \textbf{Datasets.} We conduct experiments on the public benchmarks Argoverse~2~\cite{wilson2023argoverse} and nuScenes~\cite{caesar2020nuscenes}. Argoverse~2 consists of 1,000 \SI{15}{\second} driving logs collected from six U.S. cities. Each log contains RGB images from seven \SI{20}{\hertz} cameras, along with 3D vector maps annotated at \SI{10}{\hertz}. nuScenes comprises 1,000 \SI{20}{\second} scenes captured in Boston and Singapore, providing RGB images from six \SI{12}{\hertz} cameras together with city-scale BEV vector maps annotated at \SI{2}{\hertz}. We focus on three categories of map elements: lane dividers, pedestrian crossings, and road boundaries. 
To isolate the effect of 2D-only supervision from annotation noise and ambiguity, we derive 2D labels by projecting ground-truth map elements onto multi-view images. Since nuScenes only provides BEV map elements, vertex heights are estimated from LiDAR point clouds before projection. This setup ensures a controlled supervision setting in which 3D map annotations are not used for training.

% These elements are projected onto multi-view images to obtain 2D labels. Since nuScenes only provides BEV map elements, we utilize LiDAR point clouds to estimate vertex heights before projection.
% For controlled evaluation, we derive 2D image labels by projecting ground-truth HD maps onto camera views. This setup isolates the supervision modality during training (2D labels only) while enabling quantitative comparison against 3D ground truth.

{
\setlength{\tabcolsep}{3pt}

\begin{table}[t!]
\centering
\small

\caption{\small \textbf{Comparison with state-of-the-art methods on the nuScenes validation set.} Results are reported for BEV maps within a perception range of $\SI{60}{\meter}\times\SI{30}{\meter}$ on the \emph{geographically disjoint} split~\cite{lilja2024localization}. Notation follows Table~\ref{tab:av2_results}. Results are from~\cite{lilja2024localization, lowens2025pseudomaptrainer}.}

\begin{tabular}{l|c|cccc}
\toprule
\multicolumn{1}{c|}{Method} & Map GT & AP$_{\text{div}}$ & AP$_{\text{ped}}$ & AP$_{\text{bou}}$ & mAP \\
\midrule
VectorMapNet \cite{liu2023vectormapnet}  & \ding{51} & 13.5 & 13.7 & 14.9 & 14.0 \\
MapTR \cite{liao2022maptr}               & \ding{51} & 16.0 & 14.4 & 26.7  & 19.0 \\
MapQR \cite{liu2024leveraging}           & \ding{51} & 22.2$^{*}$ & 17.4$^{*}$ & 29.9$^{*}$ & 23.2$^{*}$ \\
MapTRv2 \cite{liao2024maptrv2}           & \ding{51} & 20.9 & 26.5 & 32.6 & \textbf{26.7} \\
\midrule
MapQR \cite{liu2024leveraging} + IPM \cite{mallot1991inverse} & \ding{55} & 14.5 & 10.0 & 14.8 & 13.1 \\
PseudoMapTrainer \cite{lowens2025pseudomaptrainer}            & \ding{55} & 4.1 & 18.1 & 17.4 & 13.2 \\
MapRF (Ours)                                                  & \ding{55} & 15.9 & 11.2 & 20.9 & \textbf{16.0} \\
% estimated mAP around 16.24 to 17.4
\bottomrule
\end{tabular}
\label{tab:nus_results} 
\vspace{-5mm}
\end{table}

}

% thomas.monninger@mercedes-benz.com
% We evaluate performance using mean Average Precision (mAP) based on the Chamfer distance, following~\cite{liao2022maptr, liao2024maptrv2}. The perception range is centered on the ego vehicle, spanning $[-15.0m, 15.0m]$ (X), $[-30.0m, 30.0m]$ (Y), and $[-5.0m, 3.0m]$ (Z). A prediction is considered a true positive (TP) if its Chamfer distance to the ground truth is below a predefined threshold. We use two threshold sets: $\{0.2, 0.5, 1.0\}m$ for the hard setting and $\{0.5, 1.0, 1.5\}m$ for the easy setting, as in~\cite{ding2023pivotnet, qiao2023end}. Following~\cite{liao2024maptrv2, liu2024leveraging}, we evaluate three classes of map elements: pedestrian crossings, lane dividers, and road boundaries. The final mAP is the average across all thresholds and classes.

\vspace{0.1cm} 
\noindent \textbf{Evaluation Metric.} We follow the standard evaluation protocol used in previous works~\cite{liao2024maptrv2, liu2024leveraging} for fair comparisons. The perception range, centered on the ego vehicle, is defined as $[\SI{-15}{\meter}, \SI{15}{\meter}]$ for the X-axis, $[\SI{-30}{\meter}, \SI{30}{\meter}]$ for the Y-axis, and $[\SI{-5}{\meter}, \SI{3}{\meter}]$ for the Z-axis. We adopt the mean average precision (mAP) to evaluate the quality of vectorized maps. A candidate is considered a true positive (TP) if its Chamfer distance to the ground truth is below a given threshold. We compute the average precision (AP) at distance thresholds $\{ \SI{0.5}{\meter}, \SI{1.0}{\meter}, \SI{1.5}{\meter} \}$ and report mAP over all categories.

\vspace{0.1cm} 
\noindent \textbf{Implementation Details.} We employ ResNet-50~\cite{he2016deep} as the backbone for fair comparison with previous works~\cite{liu2023vectormapnet,liao2022maptr}. The map model follows the MapQR~\cite{liu2024leveraging} architecture and training recipe, with a BEV feature size of 200$\times$100, a 6-layer decoder, 100 instance queries, and 20 points per map element. We train the map model for 6 epochs on Argoverse~2 and 24 epochs on nuScenes with a batch size of 32, using AdamW optimizer with a learning rate of 6$\times$10$^{-4}$ and cosine annealing. The MC-NeRF samples along camera rays up to \SI{35}{\meter} at \SI{0.3}{\meter} intervals and applies a 387-dimensional (387-d) frequency positional encoding. Both neural point and sampling features are 64-d, and output instance embeddings are 8-d. Unless otherwise specified, all MLPs consist of three fully connected layers with ReLU activations. We train MC-NeRF for 2 epochs on Argoverse~2 and 8 epochs on nuScenes with a batch size of 24, using AdamW optimizer with a learning rate of 4.5$\times$10$^{-4}$ and cosine annealing. We freeze the map model during MC-NeRF training for stability. The number of self-training~rounds is set to 3. During inference, we remove the MC-NeRF for efficiency.

\subsection{Comparison with State-of-the-Art Methods}

\vspace{0.1cm} 
\noindent \textbf{Results on Argoverse~2.} As shown in Table~\ref{tab:av2_results}, our method achieves performance comparable to fully supervised baselines and outperforms weakly supervised ones. Without 3D HD map annotations during training, MapRF attains \SI{50.8}{\percent}~mAP$_1$ on BEV maps and \SI{49.5}{\percent}~mAP$_{\text{2}}$ on 3D maps. It surpasses HDMapNet~\cite{li2022hdmapnet} by \SI{32.0}{\percent}~mAP$_1$ and VectorMapNet~\cite{liu2023vectormapnet} by \SI{12.9}{\percent}~mAP$_1$ and \SI{13.7}{\percent}~mAP$_{\text{2}}$, while remaining below the top-performing fully supervised methods~\cite{liao2022maptr, liao2024maptrv2, liu2024leveraging}. Notably, MapRF reaches about \SI{75}{\percent} of the performance of the reference method MapQR~\cite{liu2024leveraging}, demonstrating its effectiveness in leveraging 2D image labels for map learning. Compared to weakly supervised baselines, MapRF further outperforms MapQR~\cite{liu2024leveraging} + IPM~\cite{mallot1991inverse} by \SI{8.4}{\percent}~mAP$_1$ and \SI{8.6}{\percent}~mAP$_{\text{2}}$. Since PseudoMapTrainer~\cite{lowens2025pseudomaptrainer} does not report results on Argoverse~2, we present a comparison on the nuScenes~dataset.

\vspace{0.1cm} 
\noindent \textbf{Results on nuScenes.} As presented in Table~\ref{tab:nus_results}, our method achieves performance comparable to fully supervised methods and outperforms weakly supervised ones. Without HD map annotations during training, MapRF attains an mAP of \SI{16.0}{\percent} for BEV map construction, surpassing VectorMapNet~\cite{liu2023vectormapnet} by \SI{2.0}{\percent}~mAP while still remaining below top-performing fully supervised methods~\cite{liao2022maptr, liao2024maptrv2, liu2024leveraging}. Compared to other weakly supervised methods, MapRF further outperforms PseudoMapTrainer~\cite{lowens2025pseudomaptrainer} by \SI{2.8}{\percent}~mAP and MapQR~\cite{liu2024leveraging} + IPM~\cite{mallot1991inverse} by \SI{2.9}{\percent}~mAP. 

\vspace{0.1cm} 
\noindent \textbf{Discussion.} Although MapRF does not reach the performance of fully supervised methods, this gap is expected since our framework does not rely on GT 3D map annotations during training. By leveraging accessible 2D image labels, MapRF substantially reduces annotation costs and improves scalability. Achieving about \SI{75}{\percent} of the performance of the fully supervised MapQR~\cite{liu2024leveraging} baseline on Argoverse~2 indicates a favorable cost–accuracy trade-off. This trade-off makes MapRF suitable for scenarios where 3D map labels are unavailable.

% While MapRF does not yet match the performance of fully supervised methods trained with dense 3D map annotations, it substantially reduces annotation costs and offers improved scalability. This trade-off makes MapRF particularly suitable for scenarios where high-quality HD maps are unavailable.

% As shown in Table~\ref{tab:nus_results}, our method achieves comparable or better performance than existing weakly supervised approaches while being trained Without 3D HD map annotations during training. 
% Specifically, MapRF attains an overall \SI{16.0}{\percent}~mAP on the nuScenes validation set, outperforming MapQR~\cite{liu2024leveraging} + IPM~\cite{mallot1991inverse} (\SI{13.1}{\percent}) and PseudoMapTrainer~\cite{lowens2025pseudomaptrainer} (\SI{13.2}{\percent}) across all categories. 
% Among individual classes, MapRF achieves \SI{15.9}{\percent}, \SI{11.2}{\percent}, and \SI{20.9}{\percent} AP for lane dividers, pedestrian crossings, and road boundaries, respectively. 
% Although fully supervised methods such as MapQR~\cite{liu2024leveraging} reach higher absolute performance (\SI{23.2}{\percent}~mAP), our method retains about \SI{70}{\percent} of the supervised baseline while requiring only 2D image labels. 
% These results demonstrate that MapRF generalizes effectively to diverse urban environments and camera configurations, validating its scalability and robustness for weakly supervised online HD map construction.

\begin{table}[t!]
\centering
\vspace{2mm}

\small
\caption{\small \textbf{Ablation study of main components.} Performance progressively improves as each component is added. MC-NeRF, ST, and MRM denote the Map-Conditioned NeRF module, the self-training process, and the Map-to-Ray Matching, respectively.}

\begin{tabular}{ccc|cccc}
\toprule
MC-NeRF & ST & MRM & AP$_{\text{div}}$ & AP$_{\text{ped}}$ & AP$_{\text{bou}}$ & mAP \\
\midrule
    &            &         & 54.4 & 29.7 & 38.6 & 40.9 \\
\ding{51} &       &        & 55.3 & 34.2 & 45.4 & 45.0 \\
          & \ding{51} &        & 51.7 & 28.1 & 36.5 & 38.7 \\
\ding{51} & \ding{51} &  & 57.5 & 38.9 & 49.7 & 48.7 \\
\ding{51} & \ding{51} & \ding{51} & 58.1 & 40.5 & 49.8 & \textbf{49.5}  \\
\bottomrule
\end{tabular}
\label{tab:ablate_main} 

\vspace{-3mm}
\end{table}

\begin{table}[t!]
\centering
\small

\caption{
\small \textbf{Ablation study of Map-Conditioned NeRF.} Introducing auxiliary timestamps consistently improves performance. Moderate temporal offsets yield the best results.}

\begin{tabular}{c|cccc}
\toprule
Auxiliary Timestamps ($\Delta t$) & AP$_{\text{div}}$ & AP$_{\text{ped}}$ & AP$_{\text{bou}}$ & mAP \\
\midrule
/                 & 54.6 & 33.5 & 46.4 & 44.8 \\
$(\SI{-1.0}{\second}, +\SI{1.0}{\second})$ & 55.1 & 37.7 & 48.1 & 47.0 \\
$(\SI{-2.0}{\second}, +\SI{2.0}{\second})$ & 58.1 & 40.5 & 49.8 & \textbf{49.5} \\
$(\SI{-3.0}{\second}, +\SI{3.0}{\second})$ & 57.4 & 39.7 & 46.4 & 47.9   \\
$(\SI{-4.0}{\second}, +\SI{4.0}{\second})$ & 57.3 & 39.1 & 44.8 & 47.0\\
\bottomrule
\end{tabular}
\vspace{-5mm}
\label{tab:ablate_nerf} 
\end{table}

\subsection{Ablation Study and Analysis} \label{sec: ablation}
We conduct ablation studies to evaluate the effectiveness of each component of MapRF. Experiments are performed on the Argoverse~2 dataset for 3D map construction.

\vspace{0.1cm} 
\noindent \textbf{Ablation of Main Components.} We analyze the contributions of the main components of MapRF. As shown in Table~\ref{tab:ablate_main}, each component progressively improves performance. The initial map model achieves \SI{40.9}{\percent}~mAP. Introducing the Map-Conditioned NeRF results in a \SI{4.1}{\percent}~mAP gain, demonstrating its effectiveness in generating geometrically accurate pseudo labels. Incorporating self-training with MC-NeRF leads to an additional \SI{3.7}{\percent}~mAP improvement through iterative refinement. The Map-to-Ray Matching provides a further \SI{0.8}{\percent}~mAP gain, indicating its role in stabilizing the self-training process. Notably, applying self-training alone does not improve performance (\SI{38.7}{\percent}~mAP), as the model reinforces its errors in the absence of geometric guidance.

\begin{figure}[t!]
  \centering
  \vspace{2mm}
  
  \hspace*{-4mm}
  \begin{subfigure}[t]{0.515\columnwidth}
    \centering
    \includegraphics[width=\linewidth]{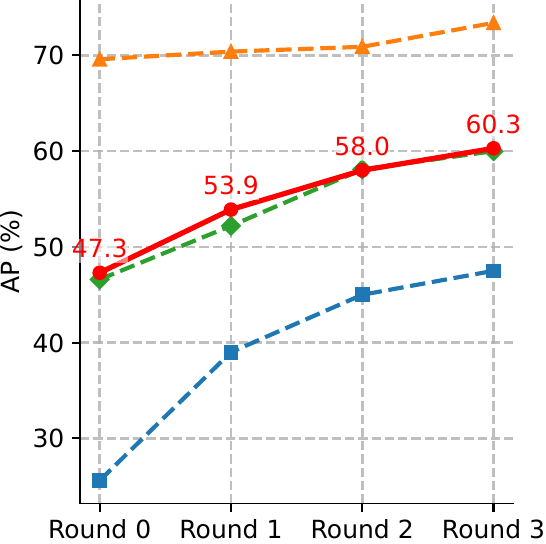}
    \caption{\footnotesize Pseudo-Label Quality}
  \end{subfigure}
  \hspace{-2mm}
  \begin{subfigure}[t]{0.515\columnwidth}
    \centering
    \includegraphics[width=\linewidth]{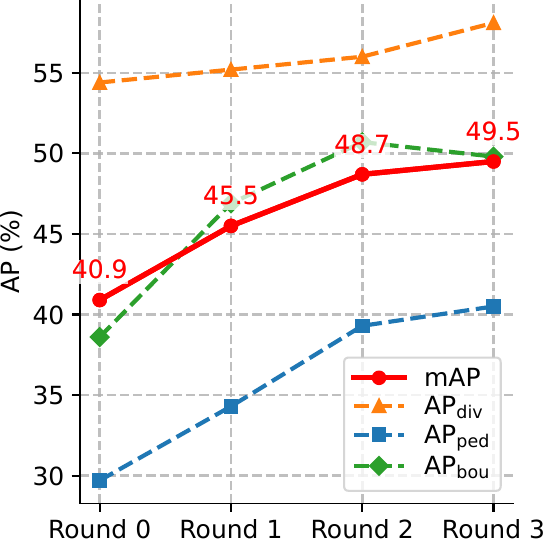}
    \caption{\footnotesize Model Performance}
  \end{subfigure}
  \caption{\small \textbf{Analysis of Self-Training Results.} Both pseudo-label quality and model performance improve progressively across self-training rounds, revealing a positive feedback loop.}
  \label{fig:self_training}
  \vspace{-2mm}
\end{figure}

\begin{table}[t!]
\centering
\small

\caption{
\small \textbf{Ablation study of Map-to-Ray Matching.} Comparison of different methods. The proposed MRM achieves the best performance with consistent improvements across all categories.}

\begin{tabular}{l|cccc}
\toprule
\multicolumn{1}{c|}{Method} & AP$_{\text{div}}$ & AP$_{\text{ped}}$ & AP$_{\text{bou}}$ & mAP \\
\midrule
Conf. Thresh. + NMS \cite{felzenszwalb2009object}  & 57.5 & 38.9 & 49.7 & 48.7  \\
Hierarchical Matching \cite{liao2024maptrv2}        & 57.7 & 39.5 & 49.5 & 48.9 \\
Map-to-Ray Matching                                & 58.1 & 40.5 & 49.8 & \textbf{49.5}       \\
\bottomrule
\end{tabular}
\label{tab:ablate_match}
\vspace{-5mm}
\end{table}

\vspace{0.1cm} 
\noindent \textbf{Ablation of Map-Conditioned NeRF.} We analyze the effect of the spatiotemporal context in the MC-NeRF. As shown in Table~\ref{tab:ablate_nerf}, introducing auxiliary rays from nearby timestamps consistently improves performance. Training without auxiliary timestamps yields \SI{44.8}{\percent}~mAP. Using temporal offsets of \SI{-2.0}{\second} and $+$\SI{2.0}{\second} achieves the best result (\SI{49.5}{\percent}~mAP), as they provide sufficient multi-view consistency for accurate geometric reconstruction. Smaller offsets (\SI{-1.0}{\second} and $+$\SI{1.0}{\second}) offer marginal gains due to limited viewpoint diversity, while larger ones ($\pm$\SI{3.0}{\second} and $\pm$\SI{4.0}{\second}) slightly degrade performance because of reduced spatial overlap across timestamps.

\begin{figure*}[t!] % The asterisk (*) makes the figure span both columns
  \centering
  \vspace{2mm}
  \includegraphics[width=0.85\textwidth]{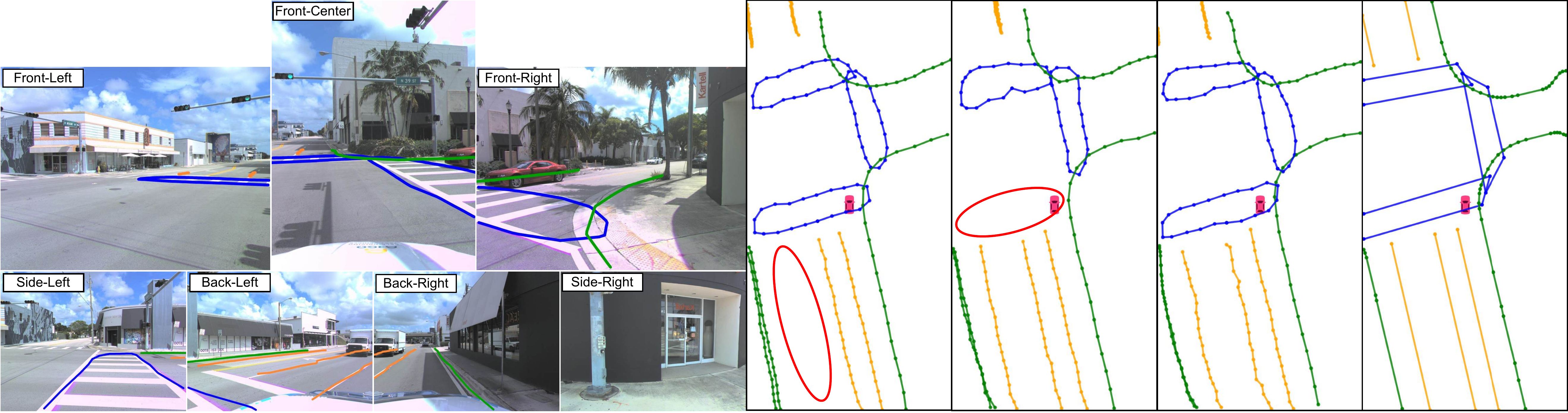}
  \includegraphics[width=0.85\textwidth]{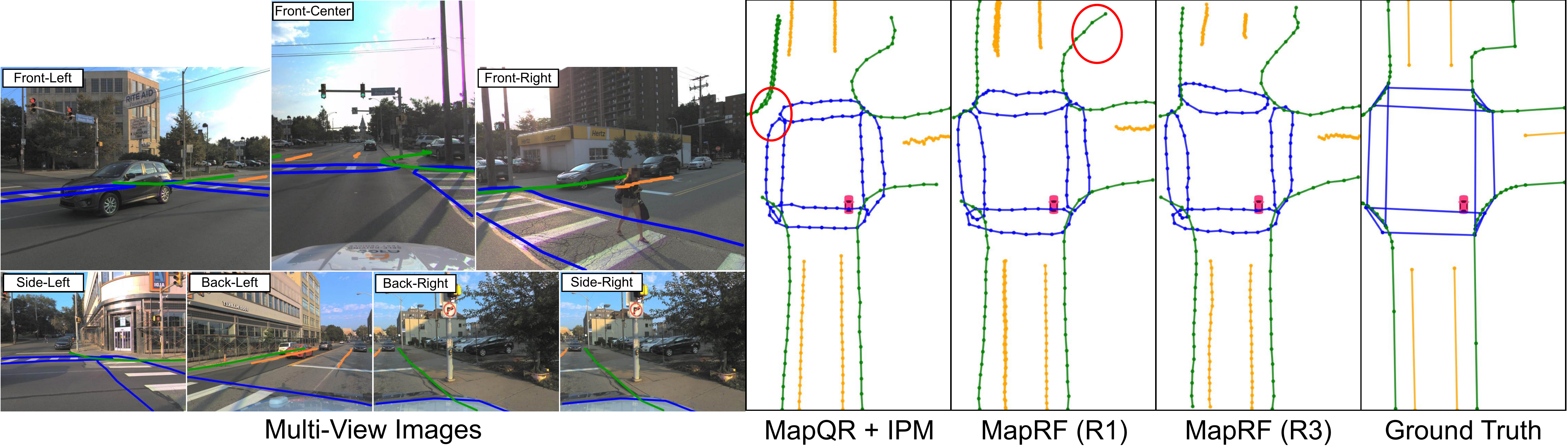}
  \caption{\small \textbf{Qualitative Results.} MapRF produces geometrically accurate maps in diverse scenes. Multi-view images are overlaid with projected predictions from MapRF (round 3). Orange, blue, and green denote lane dividers, pedestrian crossings, and road boundaries,~respectively.}
  \label{fig:qualitative}
  \vspace{-5mm}
\end{figure*}

\vspace{0.1cm} 
\noindent \textbf{Ablation of Map-to-Ray Matching.} We evaluate the proposed MRM strategy against two alternatives: (1) hierarchical matching~\cite{liao2024maptrv2}, and (2) confidence thresholding with Non-Maximum Suppression (NMS)~\cite{felzenszwalb2009object}. As shown in Table~\ref{tab:ablate_match}, MRM achieves the best overall performance at \SI{49.5}{\percent}~mAP, showing consistent improvements across all categories. This gain likely stems from its use of GT 2D annotations instead of noisy pseudo labels or model confidence, thereby reducing error propagation. Overall, MapRF remains robust to different strategies, as all methods perform around \SI{49}{\percent}~mAP.

\vspace{0.1cm} 
\noindent \textbf{Analysis of Self-Training Results.} We analyze the results of different rounds of self-training in MapRF. As shown in Fig.~\ref{fig:self_training}, both pseudo-label quality and model performance progressively improve with each self-training round. The pseudo labels become more geometrically accurate, yielding a \SI{13}{\percent}~mAP gain after three rounds, while model performance increases by \SI{8.6}{\percent}~mAP. We observe consistent improvements across all classes, with larger gains in challenging categories. These results demonstrate a positive feedback loop, where refined pseudo labels lead to better model predictions, and the improved model in turn generates higher-quality labels.

\vspace{0.1cm} 
\noindent \textbf{Qualitative Results.} We present qualitative results to illustrate the effectiveness of the proposed framework. As shown in Fig.~\ref{fig:qualitative}, MapRF produces geometrically accurate maps in diverse driving scenes and remains robust under challenging visual conditions. These results further show that predictions are progressively refined during NeRF-guided self-training.

\section{Conclusion and Future Work} \label{sec:conclusion}

In this paper, we present MapRF, a weakly supervised framework for online HD map construction that reduces the reliance on expensive map annotations. The framework leverages accessible 2D image labels and introduces a novel NeRF design, conditioned on map predictions, to generate high-quality pseudo labels for self-training. To further enhance robustness, we propose a Map-to-Ray Matching strategy that mitigates concept drift by enforcing consistent alignment between map predictions and 2D labels. Extensive experiments demonstrate the effectiveness of MapRF, achieving competitive performance while substantially reducing annotation costs. While the framework assumes the availability of 2D labels for training, future work may relax this assumption by exploring fully unsupervised approaches or leveraging foundation models to provide 2D supervision signals. In addition, exploring other 3D reconstruction methods, such as Gaussian Splatting~\cite{kerbl20233d}, may further improve the quality of pseudo map labels.

% Future directions include exploring other 3D reconstruction methods such as Gaussian Splatting~\cite{kerbl20233d} to enhance pseudo-label generation, and extending the framework to label-efficient settings with limited map annotations.



% \section*{ACKNOWLEDGMENT}

% The preferred spelling of the word ÒacknowledgmentÓ in America is without an ÒeÓ after the ÒgÓ. Avoid the stilted expression, ÒOne of us (R. B. G.) thanks . . .Ó  Instead, try ÒR. B. G. thanksÓ. Put sponsor acknowledgments in the unnumbered footnote on the first page.

% %%%%%%%%%%%%%%%%%%%%%%%%%%%%%%%%%%%%%%%%%%%%%%%%%%%%%%%%%%%%%%%%%%%%%%%%%%%%%%%%

% References are important to the reader; therefore, each citation must be complete and correct. If at all possible, references should be commonly available publications.

% 60个左右，一页
\bibliographystyle{IEEEtran}

\begingroup
\tiny
\bibliography{main}
\endgroup

% \bibliography{references}

\end{document}